\documentclass{elsart}

 \usepackage{graphicx}

\usepackage{amssymb}

\begin{document}

\begin{frontmatter}



\title{Characterization of the convergence of stationary Fokker--Planck learning}


\author{Arturo Berrones}

\address{{\small Posgrado en Ingenier\' \i a de Sistemas}\\
{\small Centro de Innovaci\'on, Investigaci\'on y Desarrollo en Ingenier\'ia
	   y Tecnolog\'ia}\\
{\small Facultad de Ingenier\' \i a Mec\'anica y El\'ectrica}\\
{Universidad Aut\'onoma de Nuevo Le\'on }\\
{\small AP 126, Cd. Universitaria, San Nicol\'as de 
los Garza, NL 66450, M\'exico}\\
{\small arturo@yalma.fime.uanl.mx}\\}

\begin{abstract}
The convergence properties of the stationary Fokker-Planck algorithm for
the estimation of the asymptotic density of stochastic search processes is studied.
Theoretical and empirical arguments for the characterization of convergence of the estimation in
the case of separable and nonseparable nonlinear optimization problems are given. Some implications
of the convergence of stationary Fokker-Planck learning for the inference of parameters in 
artificial neural network models are outlined.
\end{abstract}

\end{frontmatter}


\section{Introduction}
\label{Introduction}

The optimization of a cost function which has a number of local minima is a relevant subject in 
all fields of science and engineering. 
In particular, most of machine learning problems are stated like oftenly complex, optimization tasks \cite{bennet}.
A common setup consist in the definition of appropriate families of models that should be selected 
from data. The selection step involves the optimization of a certain cost or likelihood function, 
which is usually defined on a high dimensional parameter space. In other approaches to learning, like 
Bayesian inference \cite{neal, mackay}, the entire landscape generated by the optimization problem
associated with a set of models together with the data and the cost function is relevant.  
Other areas in which global optimization plays a prominent role
include operations research \cite{informs}, 
optimal design in engineenered systems \cite{design} and many other important applications. 

Stochastic strategies for optimization 
are essential to many of the heuristic techniques used to deal with complex, 
unstructured global optimization problems. 
Methods like simulated annealing \cite{anneal,anneal3,anneal4,suykens} and
evolutionary population based algorithms \cite{ga,eva,pelikan,hertz,suykens}, have
proven to be valuable tools, capable of give good quality solutions at a 
relatively small computational effort. 
In population based optimization,
search space is explored through the evolution of finite populations
of points. The population alternates periods of self -- adaptation, in which 
particular regions of the search space are explored in an intensive manner, and periods of
diversification in which solutions incorporate the gained information about the global
landscape. There is a large amount of evidence that indicates that some exponents
of population based algorithms are among the most efficient global optimization 
techniques in terms of computational cost and reliability. These methods, however
are purely heuristic and convergence to global optima is not
guaranteed.
Simulated annealing on the other hand, is a method that statistically assures global optimality, but 
in a limit that is very difficult to acomplish in practice. In simulated annealing a single
particle explores the solution space through a diffusive process. In order to guarantee global 
optimality, the ``temperature'' that characterize the diffusion should be lowered according
to a logarithmic schedule \cite{geman}. This condition imply very long computation times.

In this contribution the convergence properties of an
estimation procedure for the stationary density of a general 
class of stochastic search processes, recently introduced by the author \cite{berrones1}, is explored.
By the estimation procedure, promising regions of the search space can be defined on a probabilistic basis.
This information can then be used in connection with a locally adaptive stochastic or deterministic algorithm. 
Preliminary applications of this density estimation method in the improvement of nonlinear optimization
algorithms can be found in \cite{dexmont}. 
Theoretical aspects on the foundations of the method, its links to statistical mechanics and possible use of the
density estimation procedure as a general diversification mechanism are discussed in \cite{berrones2}.
In the next section we give a brief account of the basic elements of 
our stationary density estimation algorithm. Thereafter, theoretical and empirical evidence on the convergence of the density estimation is given.  
Besides global optimization, the density estimation approach may provide a novel technique for 
maximum likelihood estimation and
Bayesian inference.
This possibility, in the context of artificial neural network training,
is outlined in Section \ref{bayes}. Final conclusions and remarks are presented 
in Section \ref{conclusion}.

\section{Fokker--Planck learning of the stationary probability density of a stochastic search}

We now proceed with a brief account of the stationary density estimation procedure on which the present
work is based. Consider the minimization of a cost function of the form $V(x_1, x_2, ..., x_n, ..., x_N)$ 
with a search space defined over 
$L_{1,n} \leq x_n \leq L_{2,n}$. A stochastic search process for this problem is modeled by

\begin{eqnarray} \label{langevin}
\dot{x}_n = - \frac{\partial V}{\partial x_n} + \varepsilon(t) ,
\end{eqnarray} 

\noindent
where $\varepsilon(t)$ is an additive noise with zero mean. 
Equation (\ref{langevin}), 
known as Langevin equation in the statistical physics literature \cite{risken,vankampen}, 
captures the essential properties of a general stochastic search. In particular, the 
gradient term gives a mechanism for local adaptation, while the noise term provides a basic
diversification strategy. Equation (\ref{langevin}) can be interpreted as an overdamped
nonlinear dynamical system composed by $N$ interacting particles in the presence of additive
white noise. The stationary density estimation is based on an analogy with this
physical system, considering reflecting boundary conditions. It follows that the stationary 
conditional density for particle $n$ satisfy the linear partial differential equation,

\begin{eqnarray}\label{sfp}
D\frac{\partial p(x_n | \{ x_{j\neq n} = x_j^{*} \} )}{\partial x_n}
+p(x_n | \{ x_{j\neq n} = x_j^{*} \} ) \frac{\partial V}{\partial x_n} = 0 .
\end{eqnarray}

\noindent which is a one dimensional Fokker--Planck equation. 
An important consequence of Eq. (\ref{sfp}) is that the marginal $p(x_n)$ can be
sampled by drawing points from the conditional $p(x_n | \{ x_{j\neq n} = x_j^{*} \} )$
via a Gibbs sampling \cite{geman}. Due to the linearity of the
Fokker -- Planck equation, a particular form of Gibbs sampling can be constructed, 
such that its not only possible to sample the marginal density, but to give 
an approximate analytical expression for it.
From Eq. (\ref{sfp}) follows 
a linear second order 
differential equation for the cumulative distribution $y(x_n | \{ x_{j\neq n} = 
x_j^{*} \} ) = \int _{-\infty}^{x_n} 
p(x^{'}_{n} | \{ x_{j\neq n} = x_j^{*} \} ) dx^{'}_{n}$,

\begin{eqnarray}\label{sfpm}
\frac{d^{2}y}{dx_{n}^{2}}+\frac{1}{D}\frac{\partial V}{\partial x_n}\frac{dy}{dx_n}
= 0 ,\\ \nonumber
\\ \nonumber
y(L_{1,n})=0, \quad y(L_{2,n})=1 .
\end{eqnarray}

\noindent
The boundary conditions $y(L_{1,n})=0$, $y(L_{2,n})=1$ came from the fact that
the densities are normalized over the search space. 
Random deviates can be drawn from the density $p(x_n | \{ x_{j\neq n} = x_j^{*} \} )$ 
by the inversion method \cite{devroye}, 
based on the fact 
that $y$ is an uniformly distributed random variable
in the interval $y \in [0, 1]$. Viewed as a function of the random variable $x_n$, 
$y(x_n | \{ x_{j\neq n} \})$ can be approximated through a 
linear combination of functions from a complete set 
that satisfy the boundary conditions in the interval of interest,

\begin{eqnarray}\label{set}
\hat{y}(x_n | \{ x_{j\neq n} \})=\sum_{l=1}^{L} a_l \varphi _l  ( x_n  ) .
\end{eqnarray}

\noindent
Choosing for instance, a basis in which $\varphi _ l ( 0 ) = 0$, the $L$ coefficients
are uniquely defined by the evaluation of Eq. (\ref{sfpm}) in $L-1$ interior points. In this way, the 
approximation of $y$ is performed by solving a set of $L$ linear algebraic equations, involving
$L-1$ evaluations of the derivative of $V$. 
The basic sampling procedure, that we will call here Stationary Fokker--Planck (SFP) sampling,
is based on the iteration of the following steps:

\noindent
{\bf 1)} Fix the variables $x_{j\neq n} = x_j^{*}$ and approximate $y(x_n | \{ x_{j\neq n} \})$
 by the use of formulas
(\ref{sfpm}) and (\ref{set}).

\noindent
{\bf 2)} By the use of $\hat{y}(x_n | \{ x_{j\neq n} \})$ 
construct a lookup table in order to
generate a deviate 
$x_n^{*}$ drawn from the stationary distribution
$p(x_n | \{ x_{j\neq n} = x_j^{*} \})$.

\noindent
{\bf 3)} Update $x_n = x_n^{*}$ and repeat the procedure for a new variable $x_{j \neq n}$.

An algorithm
for the automatic learning of the equilibrium distribution of the
diffusive search process described by Eq. (\ref{langevin}) can be based on
the iteration of the three steps of the SFP sampling.
A convergent representation for $p(x_n)$ is obtained after taking the average
of the coefficients $a$'s in the expansion (\ref{set})
over the iterations. 
In order to see this, consider the
expressions for the marginal density and the conditional distribution,

\begin{eqnarray}
p(x_n) = 
\int p(x_n | \{ x_{j\neq n} \}) 
p(\{ x_{j\neq n} \}) d\{ x_{j\neq n} \} ,
\end{eqnarray}

\begin{eqnarray}
y(x_n | \{ x_{j\neq n} \}) =
\int_{-\infty}^{x_n} p(x^{'}_{n} | \{ x_{j\neq n} \} ) dx^{'}_{n} .
\end{eqnarray}

\noindent
From the last two equations follow that the marginal $y(x_n)$ is given by the 
expected value of the conditional $y(x_n | \{ x_{j\neq n} \} )$ 
over the set $\{x_{j\neq n}\}$,

\begin{eqnarray}
y(x_n) = E_{\{ x_{j\neq n} \}} [y(x_n | \{ x_{j\neq n} \} )] .
\end{eqnarray}

\noindent
All the information on the set $\{x_{j\neq n}\}$ is stored in the coefficients
of the expansion (\ref{set}). Therefore 

\begin{eqnarray}\label{setav}
\left< \hat{y} \right> =\sum_{l=1}^{L} \left< a_l \right> \varphi _l  ( x_n  ) 
\to y(x_n) ,
\end{eqnarray}

\noindent where the brackets represent the average over the iterations of the SFP sampling.

\section{Convergence of stationary Fokker--Planck learning}\label{converge}

The marginals $p(x_n) = dy(x_n)/dx_n$ give the probability that a diffusive particle be at any region 
$x_n dx_n$ inside the search interval $[L_{1,n}, L_{2,n}]$, under the action of the cost function.
Convergence of the stationary density estimation procedure depends on: 

i) The existence of the stationary state.

ii) Convergence of the SFP sampling.

Conditions for the existence of the stationary state for general multi--dimensional Fokker--Planck
equations can be found in \cite{risken}. For our particular reflecting boundary case, 
in which the cost function and the diffusion coefficient do not depend on time, the basic requirement
is the absence of singularities in the cost function.

By the evaluation of Eq. (\ref{setav}) at each iteration of a SFP sampling the stationary density
associated with the stochastic search can be estimated, and the accuracy of the estimate improves over
time. We call this procedure a Stationary Fokker--Planck Learning (SFPL) of a density. 
The convergence of the SFPL follows from the convergence 
of Gibbs sampling. It is known that under general conditions a Gibbs sampling displays geometric 
convergence \cite{roberts,canty}. Fast convergence is an important feature for the practical value of SFPL like a diversification
mechanism in optimization problems.
The rigorous study of the links between the geometric
convergence conditions (stated in \cite{roberts} as conditions on the kernel in a Markov chain) with
SFPL applied on
several classes of optimization problems, should be a relevant research topic. At this point, some numerical
experimentation on the convergence of SFPL is presented.

In what follows, the specific form of the expansion (\ref{set})

\begin{eqnarray}\label{set2}
\hat{y}=\sum_{l=1}^{L} a_l \sin \left ( 
(2l-1)\frac{\pi (x_n - L_{1,n} )}{2(L_{2,n}-L_{1,n})}
\right )
\end{eqnarray}

\noindent is used.

The estimation algorithm converges in one iteration for separable
problems. A separable function is given by a linear
combination of terms, where each term involves a single variable.
Separable problems generate
an uncoupled dynamics of the stochastic search described by Eq. (\ref{langevin}).
This behavior is illustrated by the minimization of the Michalewicz's function, a 
common test function for global optimization algorithms \cite{test}. 
The Michalewicz's function in 
a two dimensional search space
is written as

\begin{eqnarray} \label{micha}
V(x_1, x_2) = -\sin x_1 
(\sin(x_1 ^{2}/\pi))^{2m}
-\sin x_2
(\sin(2x_2 ^{2}/\pi))^{2m}
\end{eqnarray}

The search space is $0 \leq x_n \leq \pi$. 
The Michalewicz's function is interesting as a test function because
for large values of $m$ the local behavior of the function gives little
information on the location of the global minimum.
For $m=10$ the global minimum of the
two dimensional Michalewicz's function
has been estimated has $V\sim -1.89$ and is roughly located
around the point $(2.2, 1.5)$, as can be seen by plotting the
function. 

\begin{figure} 
\centering	
\includegraphics[width=.7\textwidth]{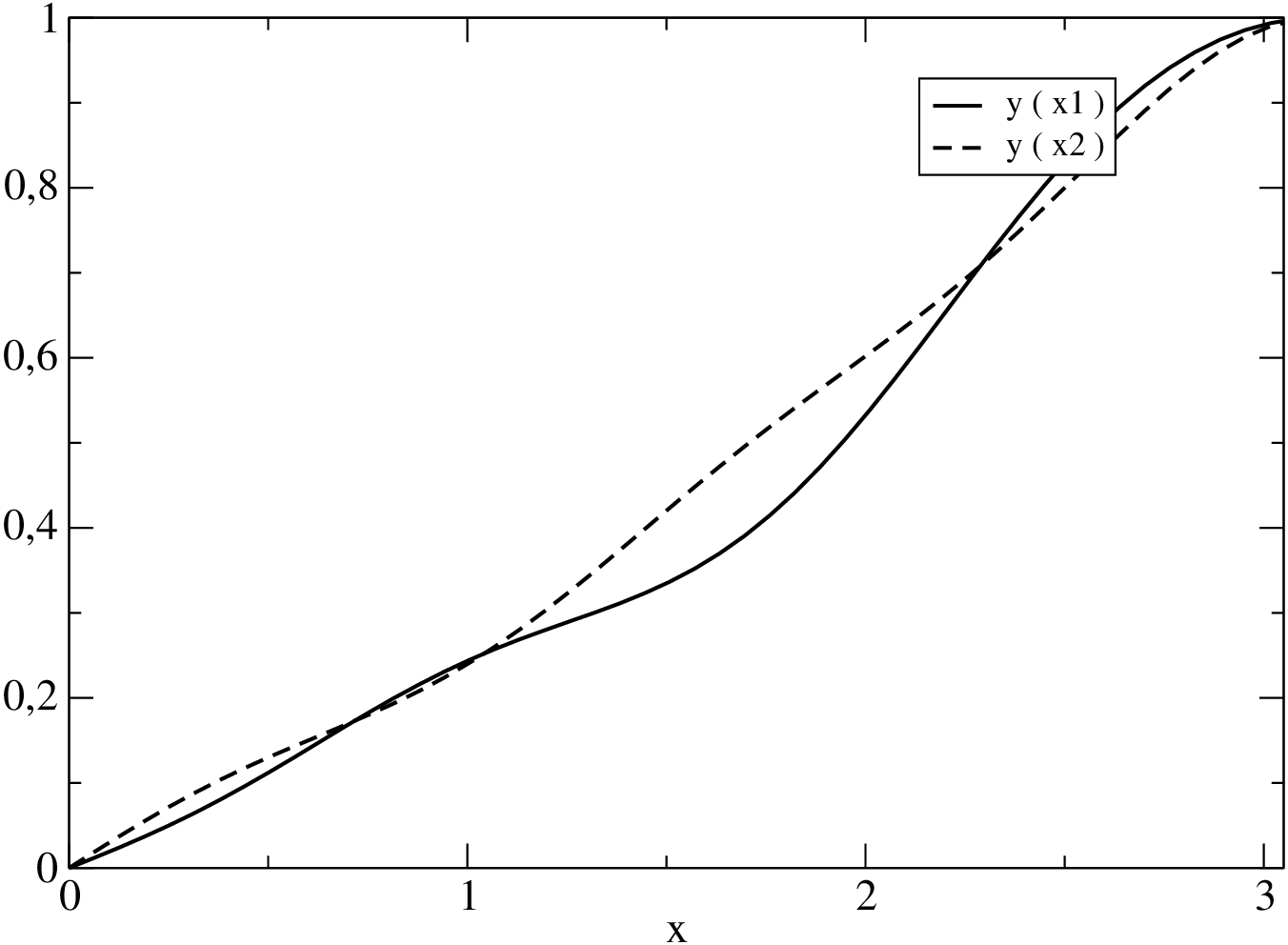}
\vskip 1.0cm
\includegraphics[width=.7\textwidth]{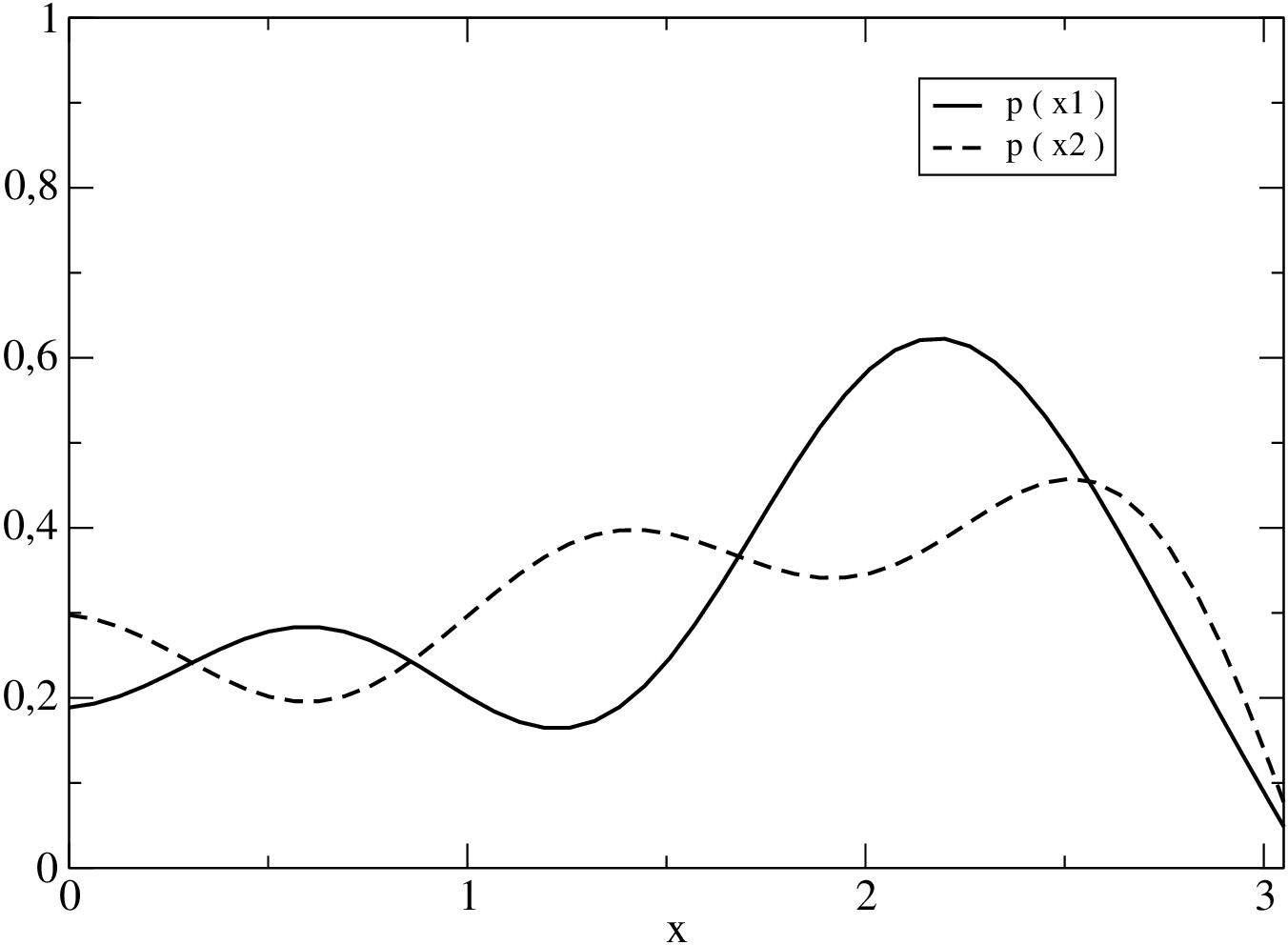}
   \caption{\label{fig1} Evaluation of
$y$ and $p$ by one iteration of the SFPL algorithm for the Michalewicz's function,
using $L=5$ and $D=1$.
Despite the very low number of gradient
evaluations used, the algorithm is capable to find a probability
structure that is consistent with the global
properties of the cost function.}
\end{figure}

\begin{figure}

\includegraphics[width=.7\textwidth]{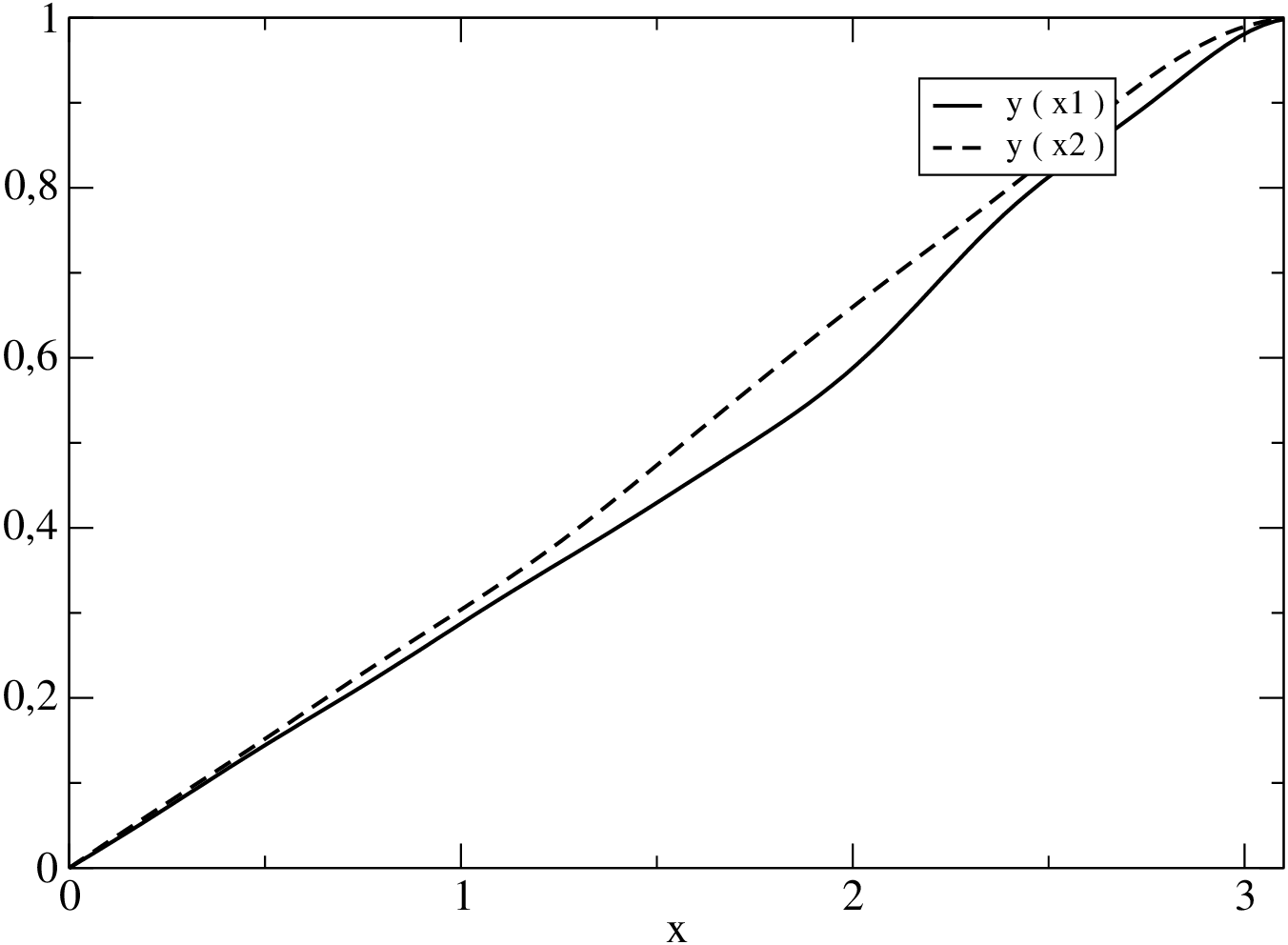}
\vskip 1.0cm
\includegraphics[width=.7\textwidth]{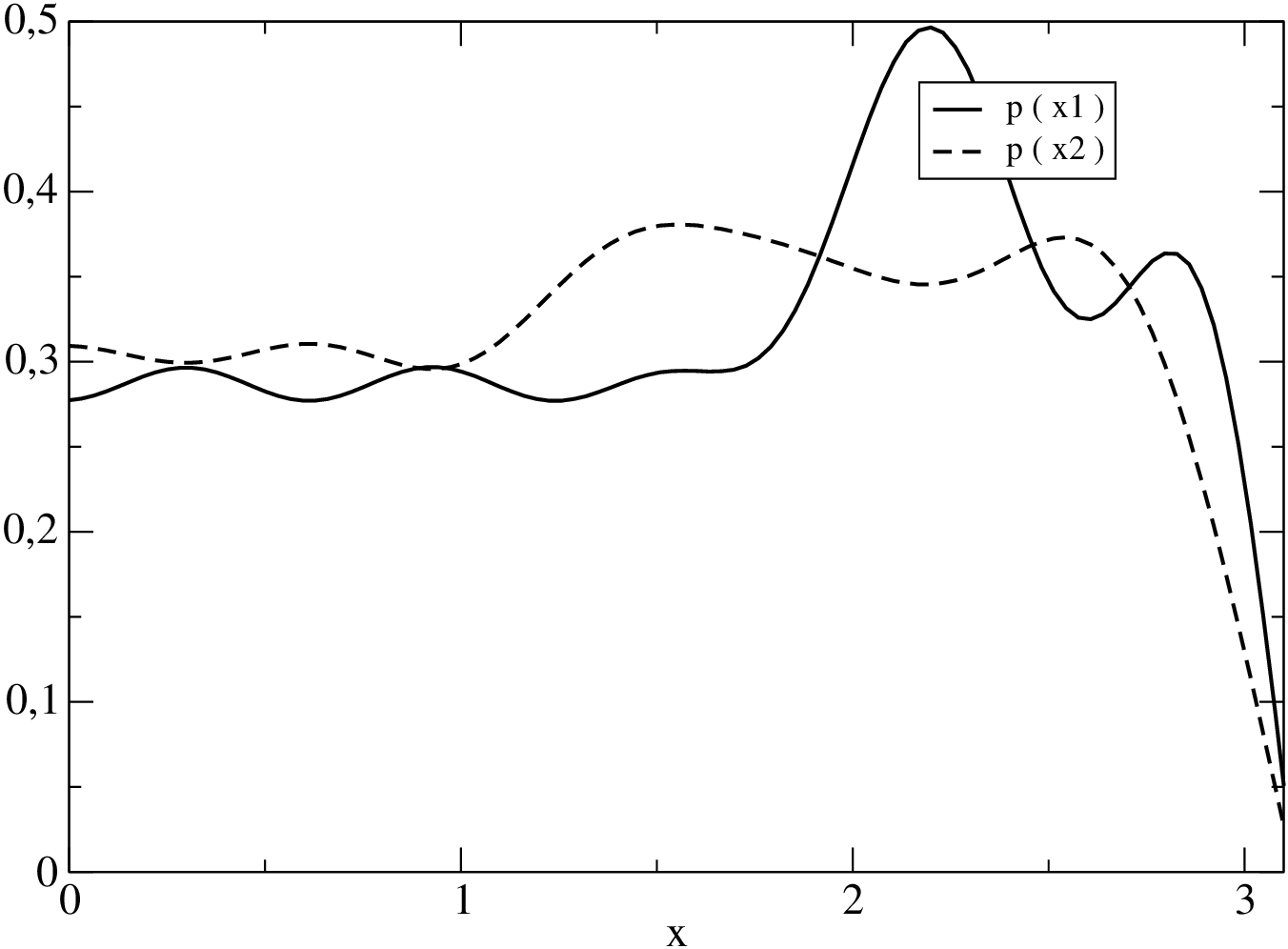}
   \caption{\label{fig2} The same case reported in Fig. (\ref{fig1}), but using
$L=10$.}
\end{figure}

\begin{figure}

\includegraphics[width=.7\textwidth]{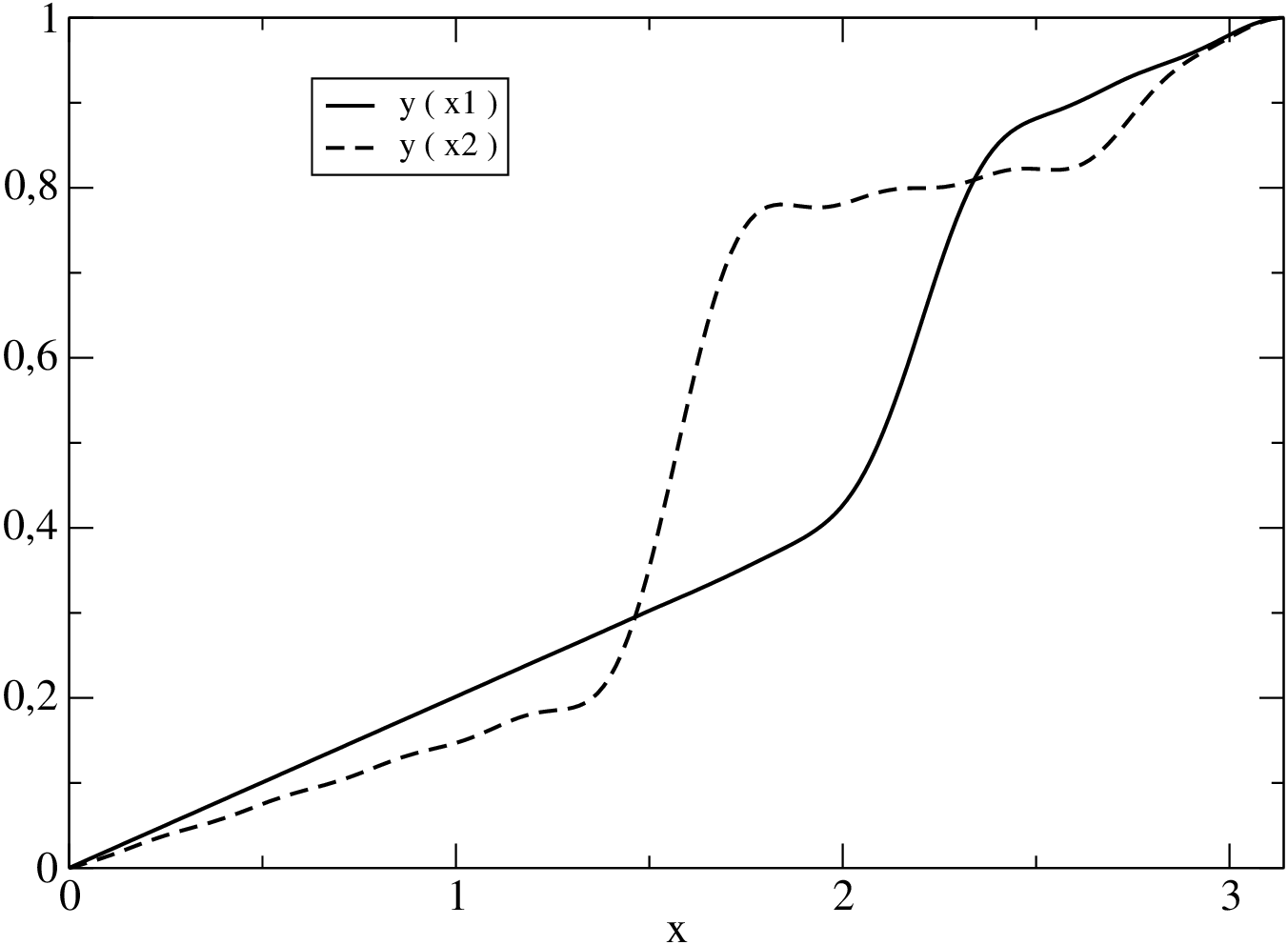}
\vskip 1.0cm
\includegraphics[width=.7\textwidth]{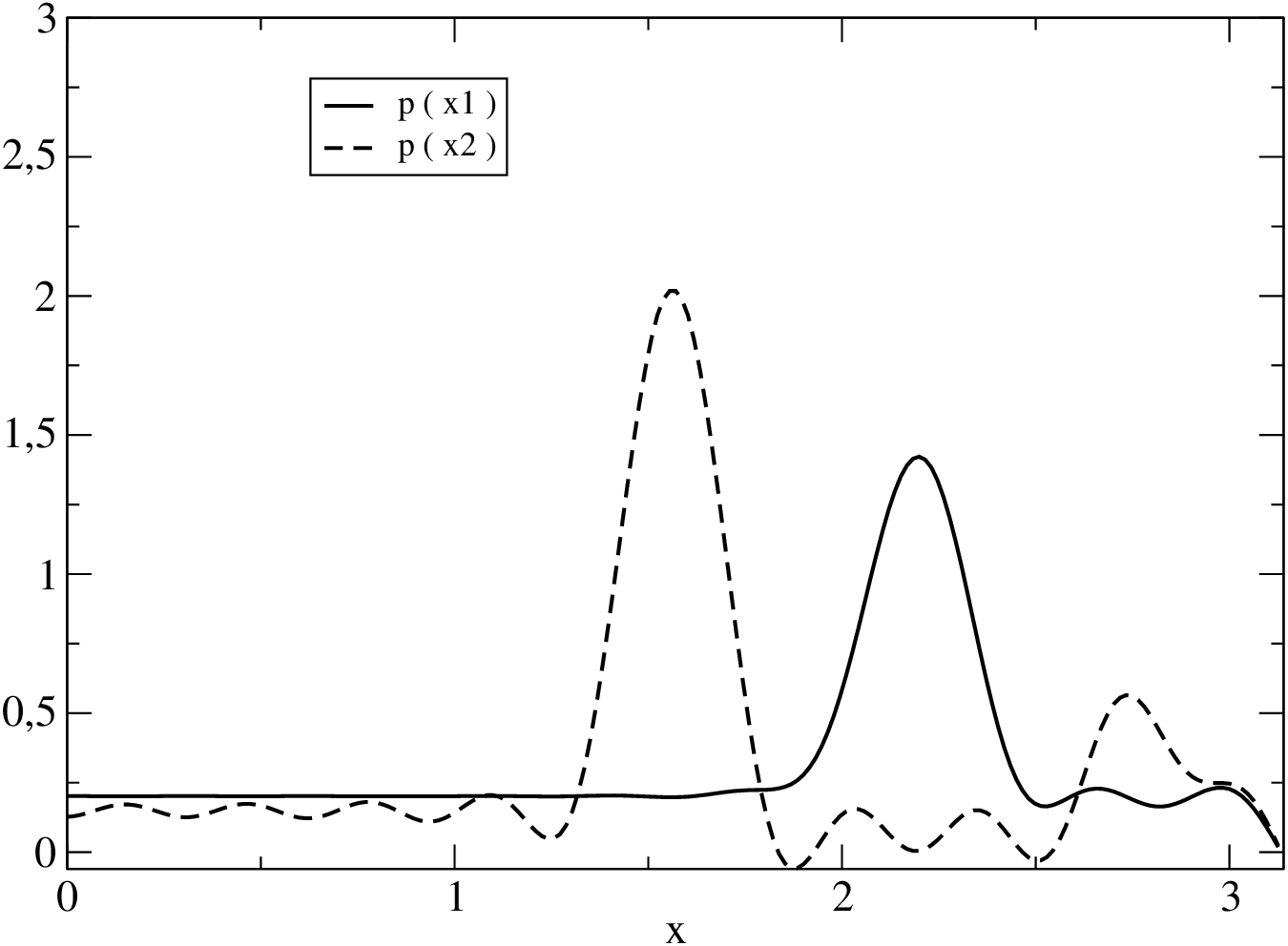}
   \caption{\label{fig3} Evaluation of
$y$ and $p$ by one iteration of the SFPL algorithm for the Michalewicz's function.
In this case $L=20$ and $D=0.4$. With
the increment in precision and the reduction of the randomness parameter,
SFPL finds a probability density that is sharply peaked around the global
minimum. Notice that the computational effort is still small, involving
only $19$ evaluations of the gradient.}
\end{figure}

The partial derivatives of function (\ref{micha}) 
with $m=10$,
have been evaluated for 
each variable at $L-1$ equidistant points separated by intervals of
size $h=\pi / L$. 
The resulting algebraic linear system has been solved by the 
LU decomposition algorithm \cite{nr}.
In Fig. (\ref{fig1}), Fig. (\ref{fig2}) and Fig. (\ref{fig3}) the functions $\hat{y}(x_1)$ and 
$\hat{y}(x_2)$ 
and their associated probability densities are shown. The densities have been estimated after a
single iteration of SFPL.
The densities 
$p(x_1)$ and $p(x_2)$ are straightforwardly calculated by taking 
the corresponding derivatives.
In Fig. (\ref{fig1}) a case with $D=1$ and $L=5$ is considered, while in
Fig. (\ref{fig2}) $D=1$ and $L=10$. In Fig. (\ref{fig3}) a smaller randomness
parameter is considered ( $D=0.4$ ), using $L=20$.
Notice that even when $D$ is high enough to allow an
approximation of $y$ with the use of very few evaluations of
the derivatives, the resulting densities 
will give populations that represent the cost function landscape
remarkably better
than those that would be obtained by uniform deviates.

The asymptotic convergence properties of the SFPL are now experimentally studied on the XOR optimization
problem, 

\begin{eqnarray}
 	\nonumber 
	f = \left\lbrace  1 + exp \left( 
			- \frac{x_7}{ 1 + exp(- x_1 - x_2 - x_5)  }
			- \frac{x_8}{ 1 + exp( - x_3 - x_4 - x_6) }
			- x_9 \right) 
			\right\rbrace ^{-2} \\ 
	\nonumber
	+ \left\lbrace 1 + exp\left( 
			- \frac{x_7}{ 1 + exp( - x_5) }
			- \frac{x_8}{ 1 + exp( - x_6) }
			- x_9 \right) 
			\right\rbrace ^{-2} \\ 
	\nonumber
	+ \left\lbrace 1 - 
				\left[ 1 +exp\left( 
					- \frac{x_7}{ 1 + exp( - x_1 - x_5 ) }
					- \frac{x_8}{ 1 + exp( - x_3 - x_6 ) }
					- x_9
					\right) 
				\right] ^{-1}
			\right\rbrace^2 \\ 
	\nonumber
	+ \left\lbrace 1 - 
				\left[ 1 + exp\left( 
					- \frac{x_7}{ 1 + exp( - x_2 - x_5 ) }
					- \frac{x_8}{ 1 + exp( - x_4 - x_6 ) }
					- x_9	
					\right) 
				\right] ^{-1}
			\right\rbrace^2 
\end{eqnarray}

\begin{figure}
	\centering
		\includegraphics[width=.7\textwidth]{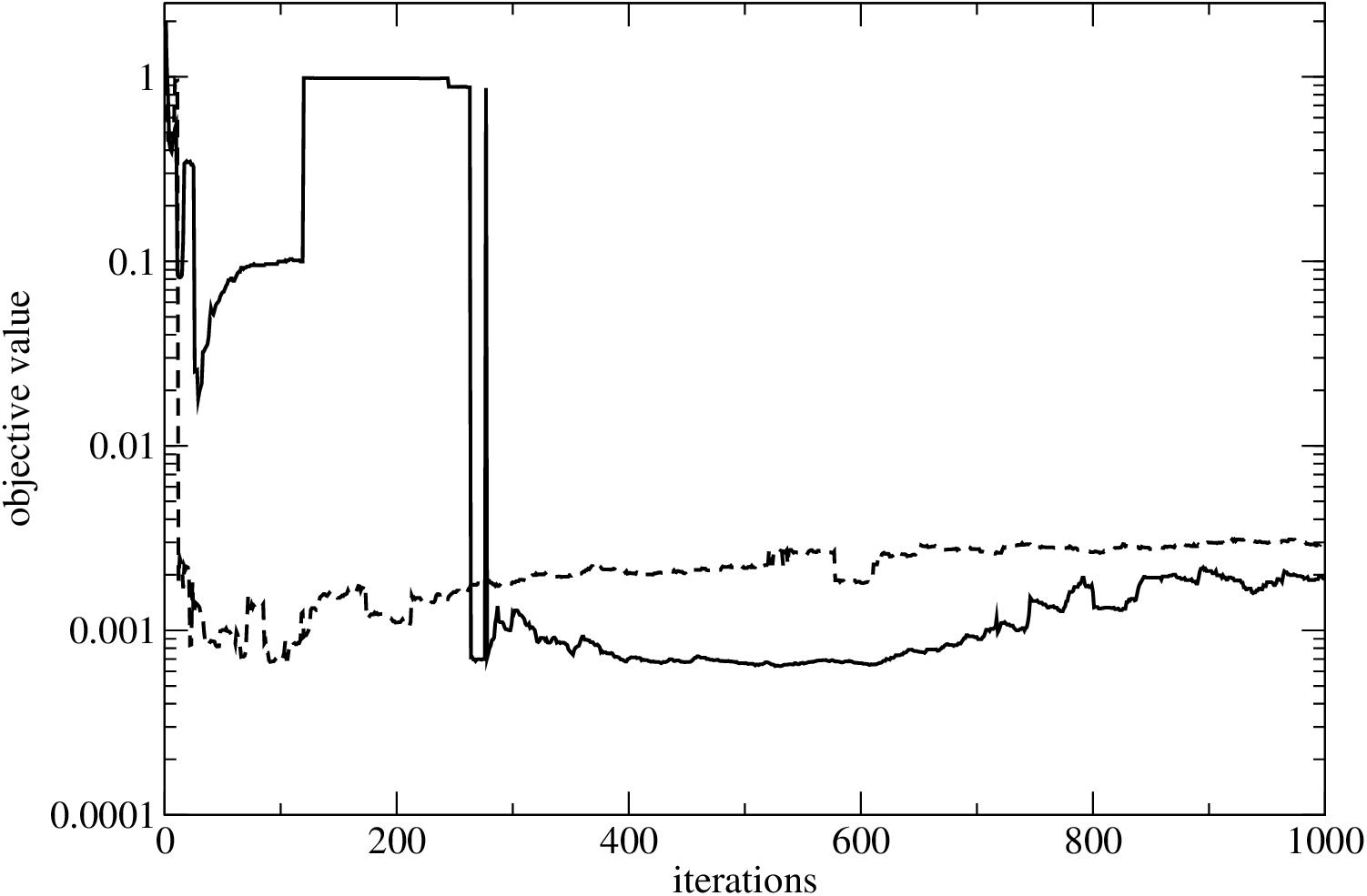}
	\caption{Objective value of the point in which the marginals of the estimated density are 
maximum for two independent trajectories.}
	\label{fig:ObXor}
\end{figure}

\begin{figure}
	\centering

		\includegraphics[width=.7\textwidth]{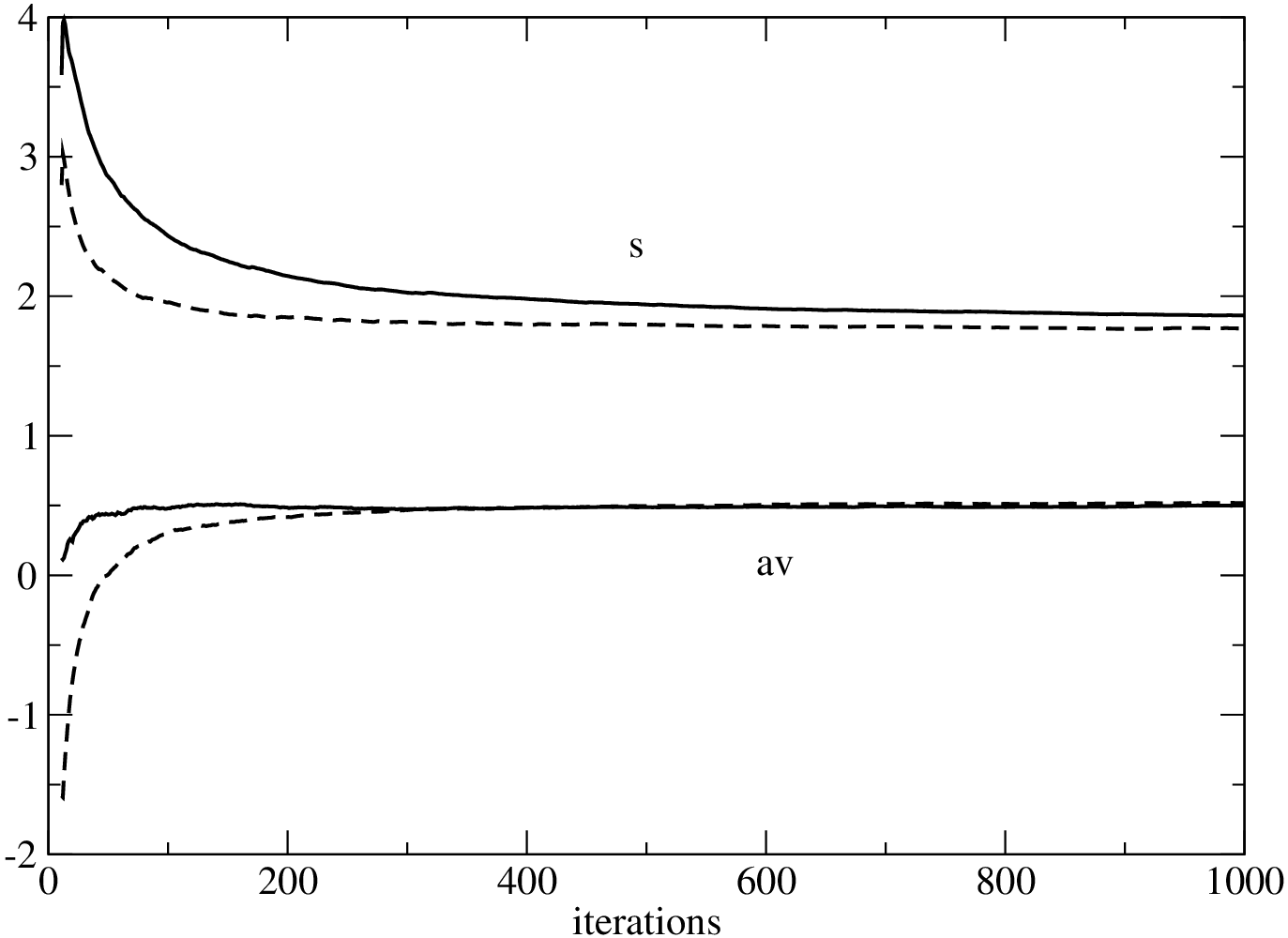}
		\vskip 1.0cm
		\includegraphics[width=.7\textwidth]{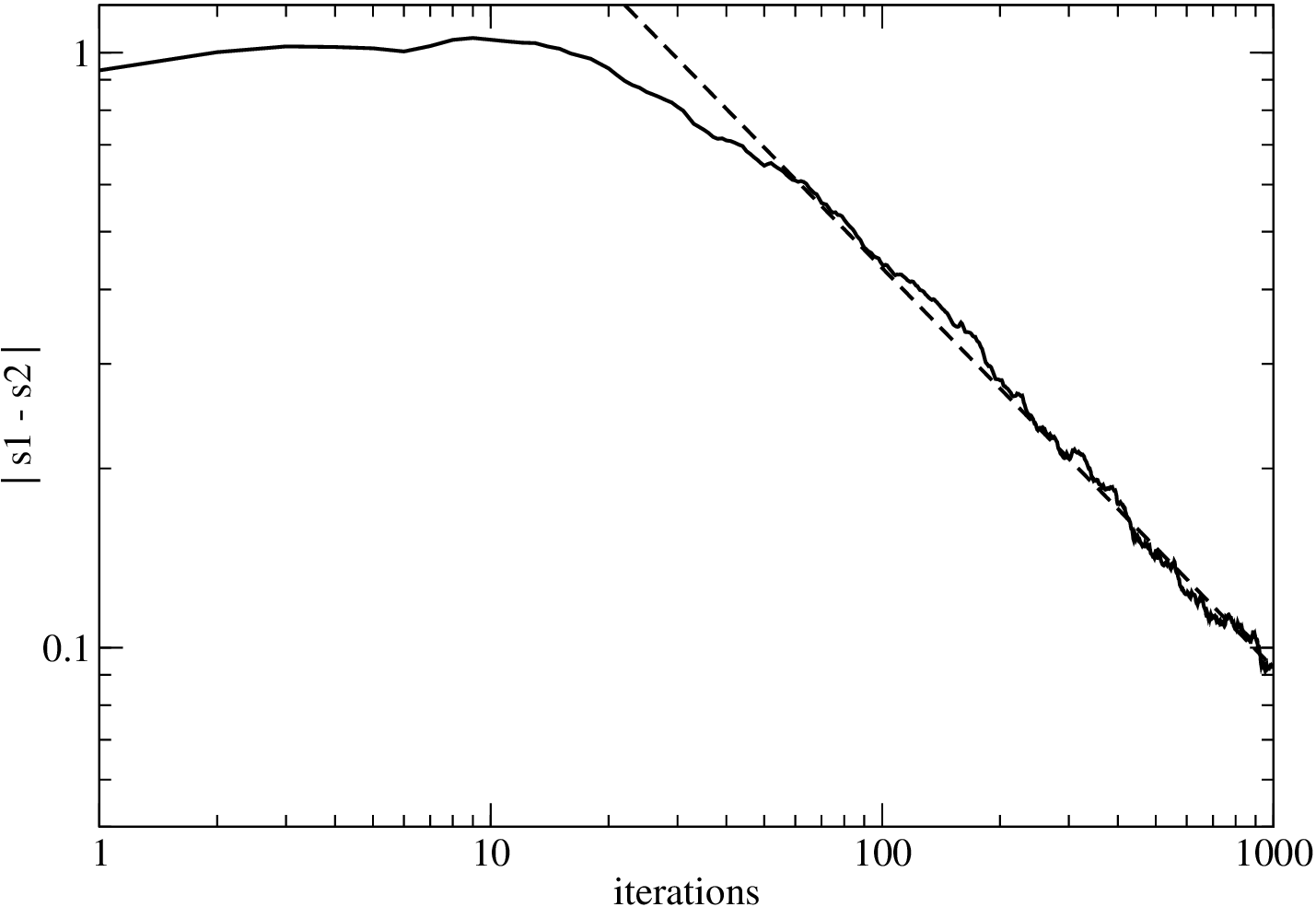}
	
	\caption{Statistical convergence of the stationary density estimation procedure on the XOR problem.
The value of the average first moment and standard deviation of the estimated marginals 
from two independent trajectories is plotted in the left. The graph on the right 
shows the distance between both average standard deviations. This distance decay 
at a geometric rate over the first $\sim 100$ iterations. Asymptotically the
distance behave like a power law 
characterized by $|s_1 - s_2| \sim M ^{-0.67}$, where $M$ is the number of iterations.  }
	\label{fig:ConXor}
\end{figure}

The XOR function is
an archetypical example that displays many of the features encountered
in the optimization tasks that arise in machine learning.
This is a case with multiple local minima \cite{pso} and strong nonlinear interactions between decision 
variables.
In the experiment reported in Fig. \ref{fig:ObXor} and Fig. \ref{fig:ConXor}, 
two independent trajectories are followed over successive
iterations. The parameters of the SFP sampler are $D = 0.01$ and $L = 200$.
In Fig. \ref{fig:ObXor} are reported the cost function values at the coordinates in which the marginals are maximum.
For each trajectory, an initial point is uniformly drawn from the search space. As can be seen, both trajectories
converge to a similarly small value of the objective function. 
The average cost function value, which is estimated by the evaluation of the cost function on $100$ points
uniformly distributed over the search space, is $2$.
After $280$ iterations, the differences between both
trajectories are around $0.05 \%$ of the average cost function value. Moreover, the differences 
in objective value of the
trajectories with respect to a putative global optimum 

\begin{eqnarray}
f(x^{*}) = 0.00026, \\ \nonumber 
x^{*}=(8.22885, -8.47952, -9.87758, 9.10184, \\ \nonumber
-4.55215, -5.05978, 9.98956, 9.96857, -4.91623),
\end{eqnarray}

\noindent is $ \leq 0.117 \%$ of the
average cost after the iteration $280$. 
The putative global optimum in the search interval has been
found by performing local search via steepest descent from a population of
points draw from the estimated density.

In order to check statistical convergence, the following measures are introduced,

\begin{eqnarray}
av = \frac{1}{N} \sum_{n=1}^{N} \left < x_n \right >, \\ \nonumber
s = \frac{1}{N} \sum_{n=1}^{N} \sqrt{\left < x_n^{2} \right > - \left < x_n \right > ^{2} },
\end{eqnarray}

\noindent where the brackets in this case represent statistical moments of the estimated 
marginals. 
Under the expansion (\ref{set2}), all the necessary integrals are easily performed analytically.

\noindent In the first graph of Fig. \ref{fig:ConXor}, 
the evolution over iterations of the SFP sampler of $s$ and $av$ for two
arbitrary and independent trajectories is plotted. A very fast convergence in the measure $av$ is evident.
The measure $s$ is further studied in the second graph of Fig. \ref{fig:ConXor}, 
where the difference on that measure among the two trajectories
is followed over iterations. 
The convergence is consistent with a geometric behavior over the first ($\sim 100$) iterations and
shows an asymptotic power law rate. 

\section{Maximum Likelihood Estimation and Bayesian Inference}\label{bayes}

Besides its applicability like a diversification strategy for local search algorithms, the fast 
convergence of the SFPL could be fruitful to give efficient aproaches to inference, for instance
in the training of neural networks.
From the point of view of statistical inference, the uncertainty about unknown parameters of a learning machine
is characterized by a posterior density for the parameters given the observed data \cite{neal,mackay}. 
The prediction of new data is then performed 
either by the maximization of this posterior (maximum likelihood estimation) or
by an ensemble average
over the posterior distribution (Bayesian inference). To be specific, suposse a {\it system}  
which generates an output $Y$ given an input $X$, such that the data is described by a distribution
with first moment $E[Y(X)] = f(X, w)$ and diagonal covariance matrix 
$\sigma ^2 I$. The problem is to estimate $f$ from a given
set of observations $S$. The parameters could be, 
for instance, different neural network weights and architectures. The observed data defines an evidence
for the different ensemble members, given by the posterior $p(w | S)$. 
In maximum likelihood estimation, training consists on finding a single set of optimal parameters that
maximize $p(w | S)$. Bayesian inference, on the other hand, is based on the fact that
the estimator of $f$ that minimizes the
expected squared error under the posterior is given by \cite{neal}

\begin{eqnarray}\label{av}
\hat{Y} = \left < f(X) \right > = \int dw f(X,w) P(w | S) ,
\end{eqnarray}

\noindent so training is done by estimating this ensemble average.

In the SFPL framework proposed here, priors are always given by uniform densities.
This choice involves very few prior assumptions, regarding  
the assignment of reasonable intervals on which the components of $w$ lie.
Under an uniform prior, and if the data present in the sample has been 
independently drawn, it turns out that the posterior is given by

\begin{eqnarray}\label{posterior}
p(w | S) \propto exp(-V/D)
\end{eqnarray}

\noindent where $D = 2\sigma ^2$ and $V$ is the given loss function. 
The SFPL algorithm can be therefore 
directly applied in order to learn the marginals $p(w_n | S)$ of the posterior (\ref{posterior}). 
By construction, these marginals will be properly normalized.

It is now argued that SFPL can be used to efficiently perform maximum likelihood and Bayesian training.
Consider again the XOR example. The associated density has been estimated
assuming a prior density  
for each parameter $w_n$ over the interval $[-10, 10]$.
The posterior density, on the other hand, is a consequence of the cost function 
given the set of training data. In Section \ref{converge} it has been shown that for 
nonseparable nonlinear cost functions like in the XOR case, SFPL converges 
to a correct estimation of the marginal densities $p(w_n)$. Therefore, the maximization of the 
likelihood is reduced to $N$ line maximizations, where $N$ is the number of weights to be estimated.
The advantage of this procedure in comparision with the direct maximization of $p(w | S)$ is evident.
On the other hand, the SFP sampler itself is designed as a generator of deviates that are drawn
from the stationary density. The average (\ref{av}) can be approximated by

\begin{eqnarray}\label{av2}
\hat{Y} = \left < f(X) \right > \approx \sum_{t} f(X,w^{(t)})  ,
\end{eqnarray}

\noindent without the need of direct simulations of the stochastic search, which is necessary in most
of other techniques \cite{neal}.

\begin{figure}
	\centering
		\includegraphics[width=.7\textwidth]{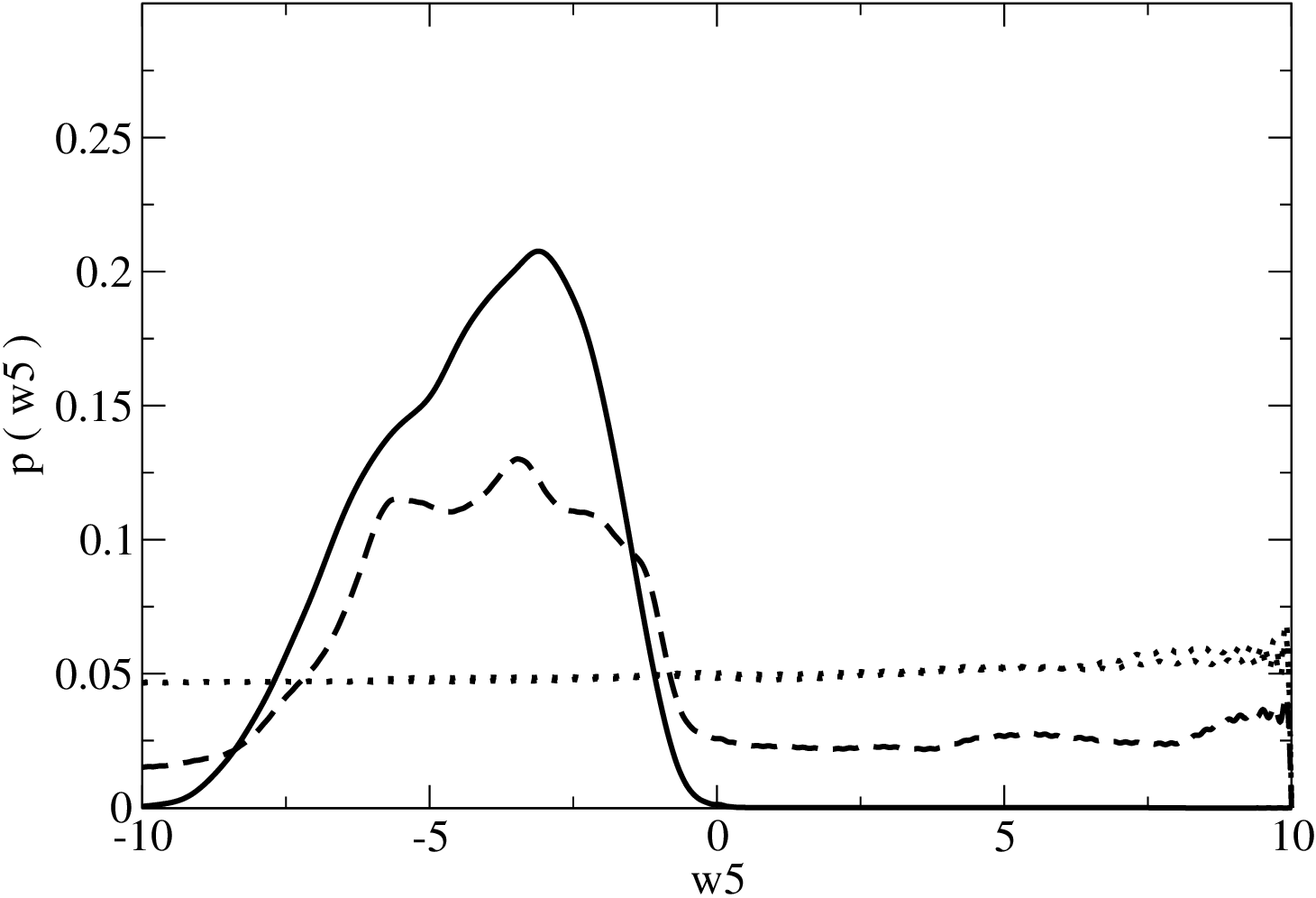}
	\caption{The probability density of a particular weight ($w_5$, a bias of one of the neurons in 
	the hidden layer) 
	of the ANN model for the XOR problem. The dotted lines correspond to cases with sample sizes of one and two.
	The dashed line is for the density that results from a sample of size three while the case for a sample
	size of four is given by the solid line.}
	\label{fig:WeightDensity}
\end{figure}

In Fig. \ref{fig:WeightDensity} 
is shown the behavior of $p(w_n)$ for a particular weight as the sample size increases. The parameters
$L = 200$ and $D = 0.01$ are fixed. The two dotted lines correspond to cases with sample sizes of one and two,
with inputs ${(0, 0)}$ and ${(0,0), (1, 1)}$
respectively. The resulting densities are almost flat in both situations. The dashed line corresponds to
a sample size of three. The sample points are ${(0, 0), (1, 1), (0,1)}$. In this case the sample is large
enough to give a sufficient evidence to favor a particular region of the parameter domain. The solid line 
corresponds to the situation in which all the four points of the data set are used for training. 
The resulting density is the sharpest. 
The parameter $D$ is proportional to the noise strength in the stochastic search. It can be selected on the basis of 
a desired computational effort, as discussed in \cite{berrones2}.
Figure \ref{fig:WeightDensity} indicates that at a fixed noise level $D$, an increase of evidence imply a decrease on
the uncertainty of the weights. This finding agrees with what is expected from the known theory of the statistical
mechanics of neural networks \cite{malzahn}, according to which the weight fluctuations decay as the
data sample grows.  

\vskip 1cm

\vskip 5cm
\begin{figure}
	\centering
		\includegraphics[width=.7\textwidth]{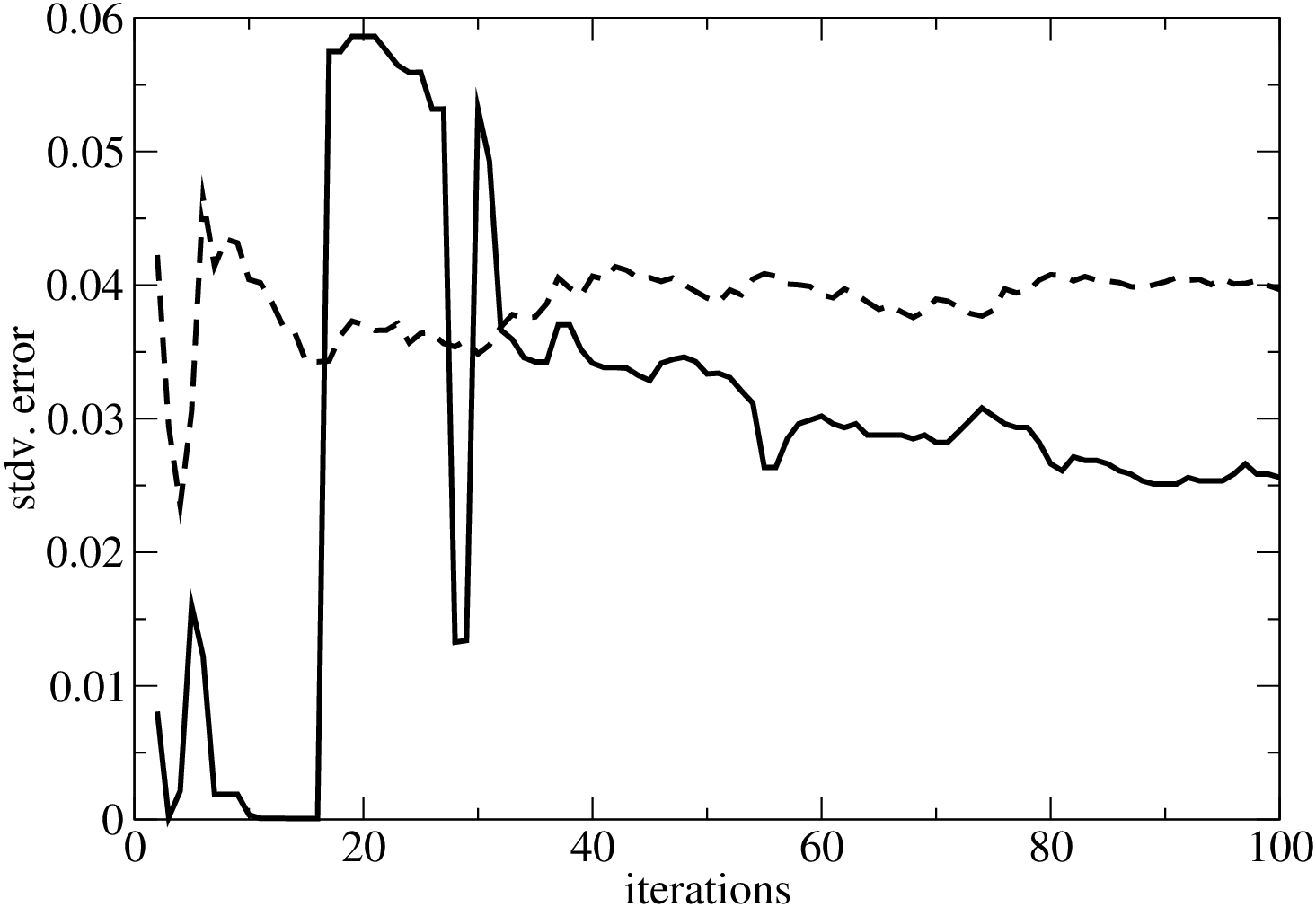}
	\caption{Performance of maximum likelihood (solid) and Bayesian (dashed) training for the XOR problem, using
	stationary Fokker--Planck learning to estimate weight distributions.}
	\label{fig:Train}
\end{figure} 

The performance of maximum likelihood and Bayesian training is reported in Fig. \ref{fig:Train}, 
using the complete sample for the inference of the weights. The standard deviation of the error 
of the networks instantiated at the inferred weigths
is reported at each iteration. The solid line corresponds to maximum likelihood training, which is essentially the same calculation already reported on Fig. \ref{fig:ObXor}.
The performance is very similar for Bayesian training, which corresponds to the dashed line. The estimation of the average (\ref{av2}) as been performed by evaluating the neural network on a weight vector drawn by the SFP sampler at each iteration. In this way, the number of terms in the sum of Eq. (\ref{av2}) is equal to the number of iterations 
of the SFP sampler.

\vskip 5cm
\begin{figure}
	\centering
		\includegraphics[width=.7\textwidth]{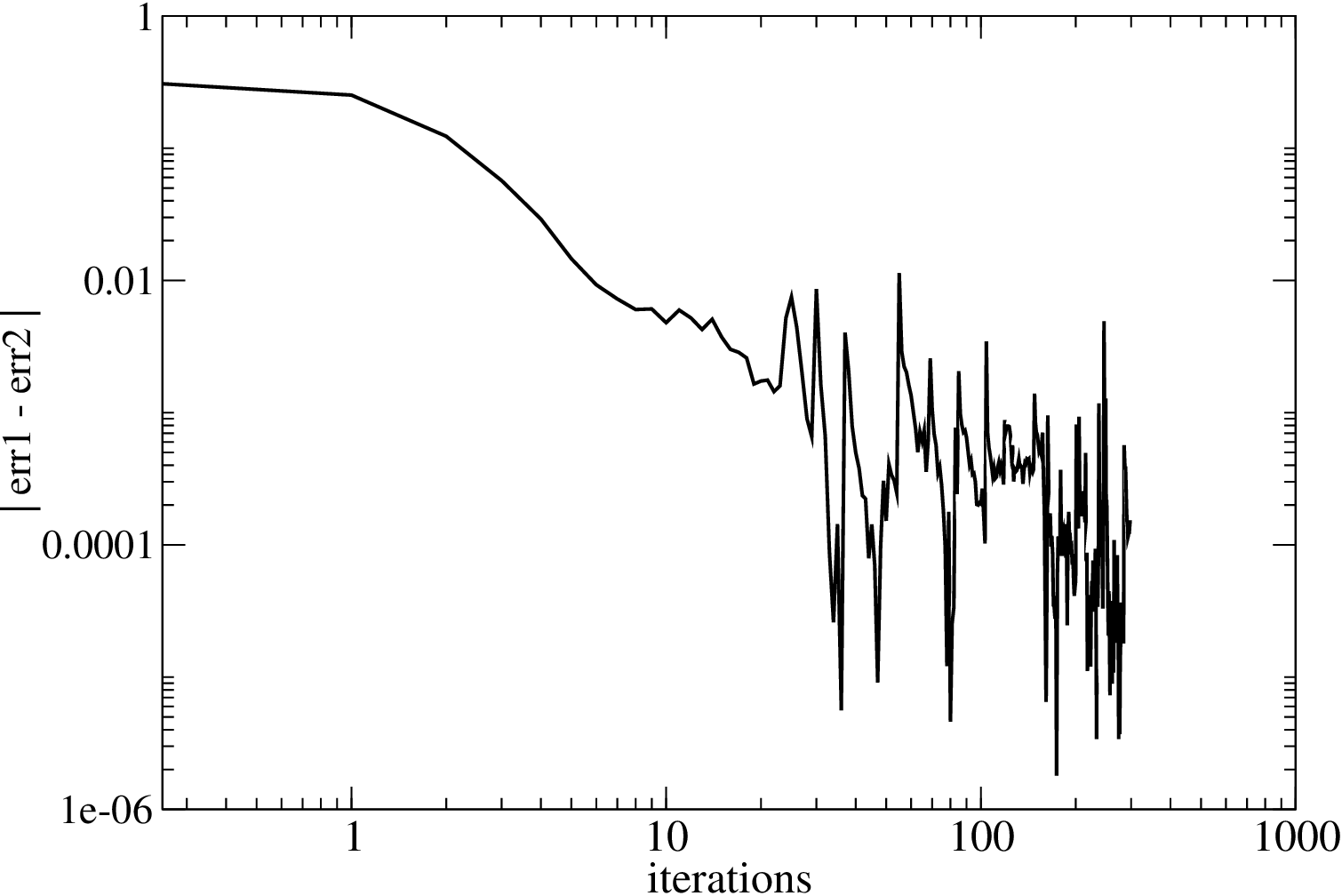}
	\caption{The difference in average cost function value for two independent trajectories in the
		 robot arm problem. The ANN as been trained using $200$ sample points.}
	\label{fig:robot}
\end{figure}

A larger example involving noisy data is now presented. Consider the ``robot arm problem'', a benchmark that has 
already been used in the context of Bayesian inference \cite{neal}. The data set is generated by the 
following dynamical model for a robot arm

\begin{eqnarray}\label{robot}
y_1 = 2.0 cos(x_1) + 1.3 cos(x_1 + x_2) + e_1 \\ \nonumber
y_2 = 2.0 sin(x_1) + 1.3 sin(x_1 + x_2) + e_2
\end{eqnarray}

\begin{figure}
	\centering
		\includegraphics[width=.7\textwidth]{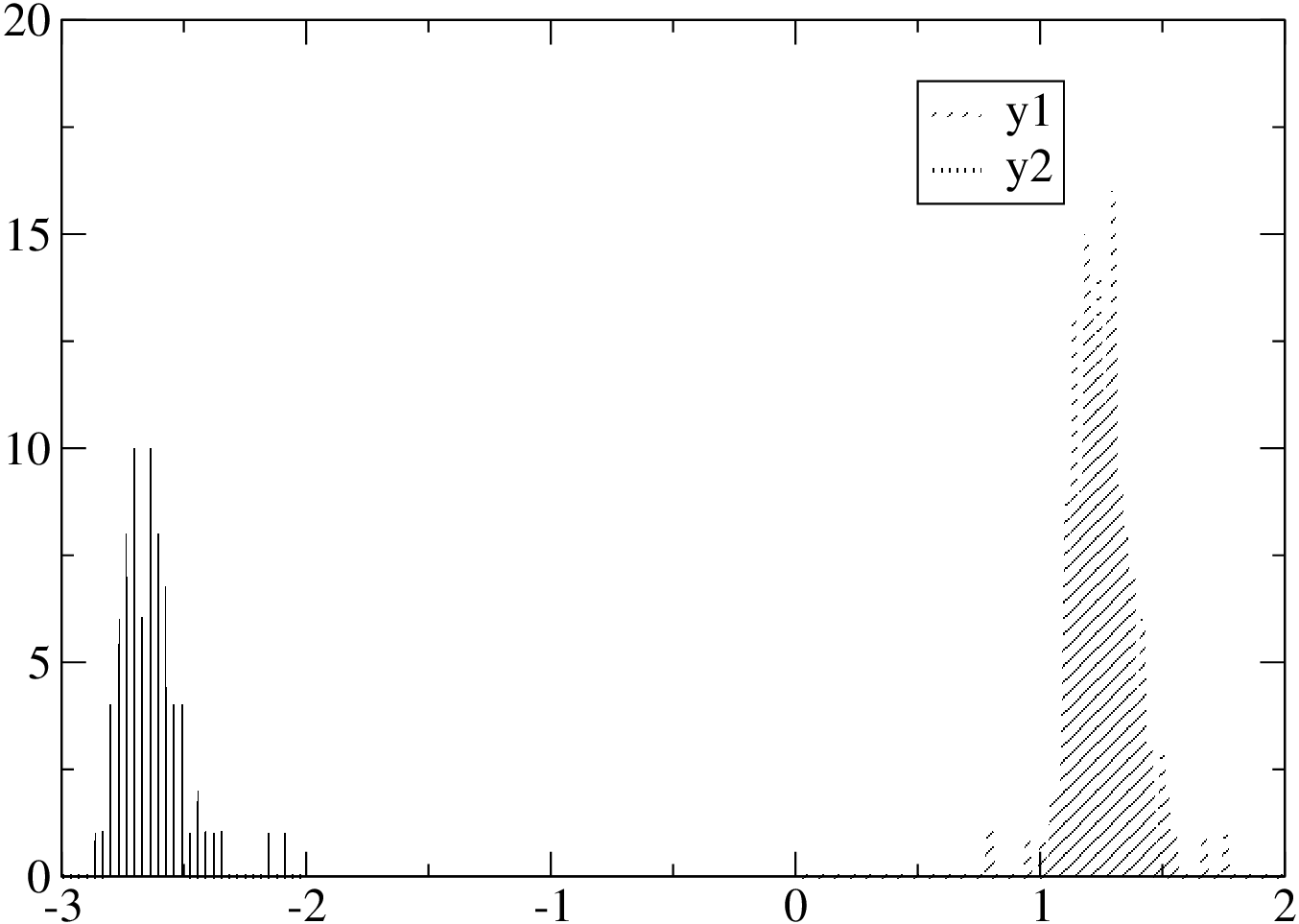}
	\caption{Bayesian predictions for the robot arm problem in the test input $(-1.471, 0.752)$.
		The SFPL parameters are $L = 300$, $D = 0.00125$ and $M = 300$, using $200$ sample points.}
	\label{fig:histogram1}
\end{figure}

\begin{figure}
	\centering
		\includegraphics[width=.7\textwidth]{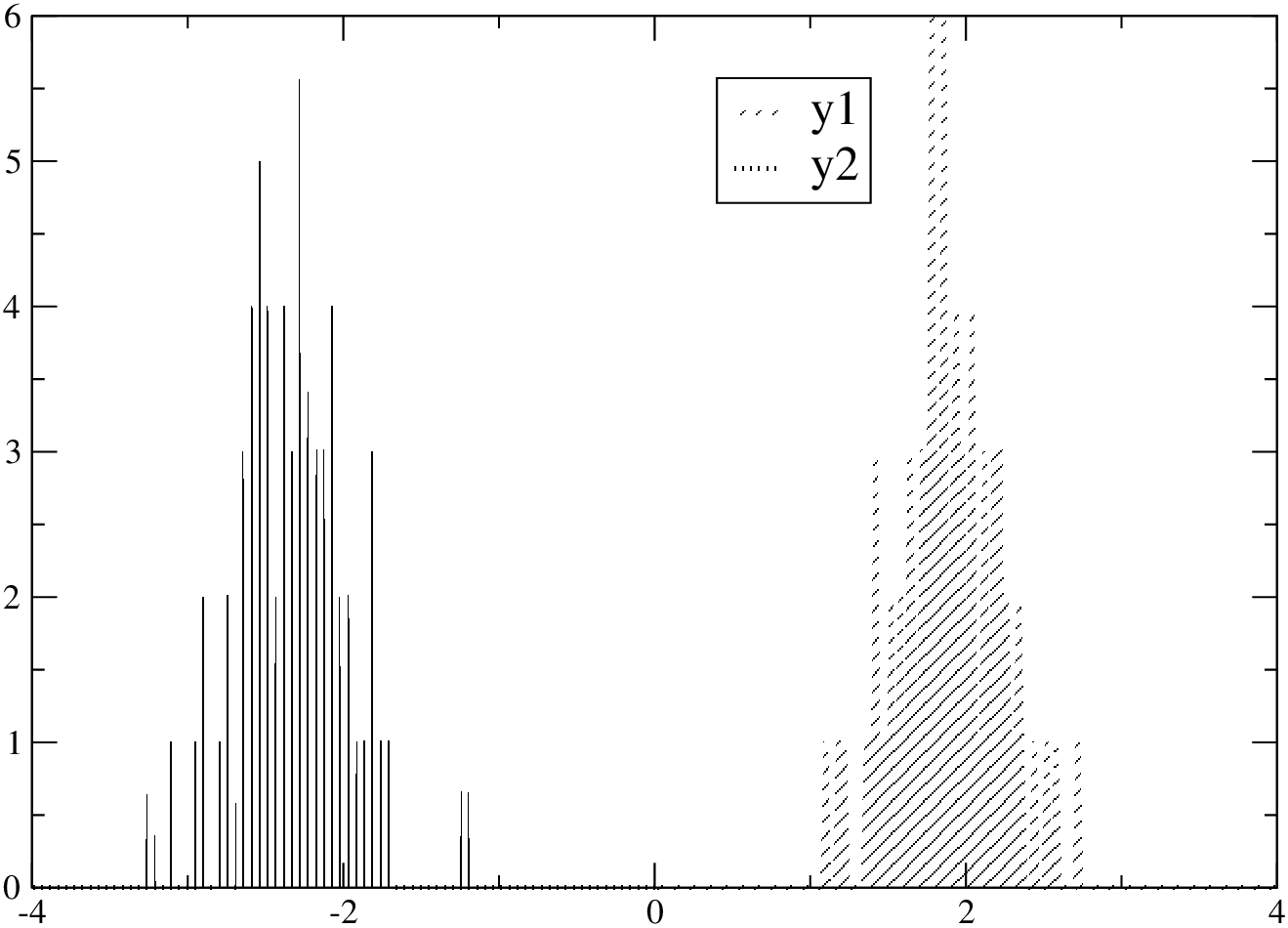}
	\caption{Same case as 
		in Fig. \ref{fig:histogram1} but with SFPL parameters given by $L=200$, $D=0.01$, $M=300$ and
		sample size of $50$ points.}
	\label{fig:histogram2}
\end{figure}

The inputs $x_1$ and $x_2$ represent joint angles while the outputs $y_1$ and $y_2$ give the resulting arm 
positions.
Following the experimental setup proposed in \cite{neal}, the inputs are uniformly distributed in the intervals
$x_1 \in [-1.932, -0.453] \cup [0.453, 1.932]$, $x_2 \in [0.534, 3.142]$.
The noise terms $e_1$ and $e_2$ are 
Gaussian and white with standard deviation of $0.1$. A sample of $200$ points is 
generated using these prescriptions. A neural network with one hidden layer consisting on $16$ hyperbolic
tangent activation 
functions is trained on the generated sample using SFP learning, considering a squared error loss function. 
The same priors are assigned to all of the weights: uniform distributions in the interval
$[-1, 1]$.
The average absolute difference in
training errors for two independent
trajectories is shown on Fig. \ref{fig:robot} for a case in which $L = 300$, $D = 0.00125$ 
and $M = 300$ iterations. 
Taking into account that the expected equilibrium square error is $err \approx D/2$, it turns out that
the differences between both trajectories are of the same order of magnitude as the expected equilibrium
error in about $10$ iterations. 
During the course of the total of $300$ iterations of SFP
of one of the trajectories, 
the network as been evaluated in the test input $(-1.471, 0.752)$. 
The resulting Bayesian prediction is shown on Fig. \ref{fig:histogram1} in
the form of a pair of histograms. 
The output that would be given by the exact model (\ref{robot}) in the absence of noise is
$(1.177, -2.847)$. The Bayesian prediction given by SFPL has its mean at $(1.24, -2.64)$ with a 
standard deviation of $\approx 0.12$ for each variable. 
Therefore the output given by the underlying model is contained in the $95 \%$ confidence interval around the Bayes
expectation. Consistent predictions can also be obtained under less precision and data. 
In Fig. \ref{fig:histogram2} are
shown the histograms obtained for a case with a sample size of $50$ points, $L=200$ and $D=0.01$. Altough the 
Bayes prediction is more uncertain, it still is statistically consistent with the underlying process.

\vskip 5cm

In the robot arm example the number of gradient evaluations needed to 
approximately converge to the equilibrium density in the $L=300$ case
was about $2(L-1)M \sim 5980$. This seems competitive
with respect to previous approaches, like the hybrid Monte Carlo strategy introduced by Neal.
The reader is referred to Neal's book \cite{neal} in order to find a very detailed application
of hybrid Monte Carlo to the robot arm problem. 
An additional advantage 
of the SFPL method lies on the fact that explicit expressions for the parameter's
densities are obtained. 

Much more detailed experimentation is under current development.
Additional studies
regarding issues like generalization of more complex ANN models under a limited amount of data, in the 
spirit of the general framework for Bayesian learning \cite{neal,mackay}, is currently a work in progress by the author.

\section{Conclusion}\label{conclusion}

Theoretical and empirical evidence for the characterization of the convergence of the density estimation of 
stochastic search processes by the method of stationary Fokker--Planck learning as been presented. 
In the context of nonlinear optimization problems, the procedure
turns out to converge in one iteration for separable problems and displays fast
convergence for nonseparable cost functions. 
The possible applications of stationary Fokker--Planck learning in
the development of efficient and reliable maximum likelihood and Bayesian ANN training techniques have been outlined.

\section*{Acknowledgement} 

This work was partially supported by the National Council of Science and Technology of
Mexico under grant CONACYT J45702-A.

\end{document}